\documentclass[10pt,twocolumn]{ICCAS}      % Use this line for a4 paper

\usepackage{comment} % Needed to meet printer requirements.
\usepackage{amsmath,amssymb,amsfonts}
\usepackage{algorithm}
\usepackage{svg}
\usepackage{algpseudocode}
\usepackage{multirow}
\usepackage{xcolor}
\usepackage{tabularx}
\usepackage{fancyvrb}
\usepackage{pdflscape}
\usepackage{lscape}
\usepackage{array}
\usepackage[caption=false,font=normalsize,labelfont=sf,textfont=sf]{subfig}
\usepackage{textcomp}
\usepackage{stfloats}
\usepackage{url}
\newcommand{\TODO}[1]{\textbf{??}}
\usepackage{verbatim}
\usepackage{graphicx}
\usepackage{cite}
\usepackage{subcaption}
\usepackage{caption}

\usepackage{amssymb}
\usepackage{orcidlink}
\usepackage{algorithm}
\usepackage{algpseudocode}
\usepackage{booktabs}
\usepackage{tikz}
\usetikzlibrary{arrows.meta, positioning, fit, backgrounds, calc, shapes.geometric, patterns, decorations.pathreplacing}

\begin{document}

\title{WaveSync: Constrained Wavefront Optimization for Synchronized Co-Speech Gestures in Humanoid Robots}

\author{ $^\dagger$Thang Tran Viet\textsuperscript{1}~$\orcidlink{0009-0003-4442-9271}$, $^\dagger$Thanh Nguyen Canh\textsuperscript{1,2}~$\orcidlink{0000-0001-6332-1002}$, Gia Huy Uong\textsuperscript{1}, Phuc Van Dinh\textsuperscript{1}, Tan Viet Tuyen Nguyen\textsuperscript{3}~$\orcidlink{0000-0001-8000-6485}$, Xiem HoangVan\textsuperscript{1,*}~$\orcidlink{0000-0002-7524-6529}$, and Nak Young Chong\textsuperscript{2,4, *}~$\orcidlink{0000-0001-5736-0769}$}

\affils{\textsuperscript{1}University of Engineering and Technology, Vietnam National University, 10000, Hanoi, Vietnam.  {\small$^\dagger$Equal contribution}\\ 
\textsuperscript{2}School of Information Science, Japan Advanced Institute of Science and Technology, Nomi, 923-1211, Ishikawa, Japan. \\ 
% (\texttt{\{thanhnc, nakyoung\}@jaist.ac.jp}) \\
\textsuperscript{3}{School of Electronics and Computer Science, University of Southampton, SO17 1BJ Southampton, United Kingdom.}\\
% (\texttt{\{23020773, 23020756, xiemhoang\}@vnu.edu.vn}) \\
\textsuperscript{4}Department of Robotics, Hanyang University, Ansan, 15588, Gyeonggi, Korea.
{\small${}^{*}$ Corresponding author}}

%%%%%%%%%%%%%%%%%%%%%%%%%%%%%%%%%%%%%%%%%%%%%%%%%%%%%%%%%%%%%%%%%%%%%%%%%%%%%%%%

\abstract{
Expressive co-speech gestures are crucial for natural human--robot interaction, yet generating them on physical humanoid robots remains challenging because, unlike virtual avatars, robots must synchronize gestures with speech under strict kinematic and actuator constraints. We present \textbf{WaveSync}, a hybrid framework in which a Large Language Model decomposes dialogue responses into structured semantic schemas and assigns per-word importance weights, forming a continuous Semantic Importance Wave. Gesture trajectories are shaped through Dynamic Movement Primitives to ensure kinematic feasibility while enhancing expressiveness. A Wavefront Optimization stage aligns gesture stroke peaks with speech emphasis peaks and resolves residual temporal conflicts through gesture-duration compression and forward propagation. Experimental evaluation across five dialogue scenarios demonstrates effective gesture--speech alignment and favorable performance in both objective and subjective evaluations. The results further show that the key components of WaveSync contribute to producing gestures that are expressive, semantically grounded, and kinematically feasible. The code, resources, and videos are available at \href{https://github.com/pairs-lab/WaveSync}{WaveSync}.
}


\maketitle

%%%%%%%%%%%%%%%%%%%%%%%%%%%%%%%%%%%%%%%%%%%%%%%%%%%%%%%%%%%%%%%%%%%%%%%%%%%%%%%%
\section{Introduction}

In Human–Robot Interaction (HRI), expressive body language is essential to make robotic communication feel natural and engaging \cite{tuyen2020conditional, canh2025efficient}. Previous studies suggest that human-like gestures and appropriate kinematic cues can enhance social presence and user trust \cite{nguyen2025context}. Successful applications of expressive body language have been demonstrated on humanoid robots \cite{yoon2019robots, tuyen2020learning, yang2026bridging}. On these platforms, nonverbal signals such as posture, gesture, and movement timing help users understand a robot's intent and build natural interactions \cite{nyatsanga2023comprehensive, tuyen2020learning2}.

Generating expressive motion usually involves three stages: defining gestures, selecting appropriate actions from the conversational context, and synchronizing those actions with speech. Researchers have proposed three main approaches to address this pipeline. Rule-based systems \cite{hu2025elegnt,salem2013closing} are generally stable but lack flexibility. They rely on predefined gesture templates and have limited ability to adapt to changing conversational cues or prosody. Data-driven models \cite{nguyen2025context,yoon2019robots} can increase gesture diversity. However, they may suffer from limited coverage of emphatic gestures in the training data and can generate trajectories that violate physical or kinematic constraints. Finally, Large Language Model (LLM)-based planners \cite{huang2025emotion} provide strong semantic grounding. However, they may generate infeasible motion specifications and introduce latency that limits real-time responsiveness \cite{chen2026real}. Across these approaches, however, speech–gesture synchronization remains a critical yet relatively underexplored challenge.

Existing state-of-the-art co-speech gesture generation methods mainly focus on virtual avatars \cite{qi2025cocogesture,zhang2024semantic}. However, physical robots are fundamentally different from avatars. While avatars can execute motions without physical actuator constraints, humanoid robots operate under strict structural and motor velocity limits. They cannot execute the rapid, high-frequency motions that avatars can easily perform. Meanwhile, the few studies deployed on physical platforms rely on simple duration matching, such as stretching the audio or adjusting keyframe speed so that both streams end at approximately the same time \cite{yoon2019robots}. This approach ignores sub-utterance temporal structure. As a result, speech emphasis peaks may not align with gesture stroke peaks, weakening speech–gesture synchronization.

To address this gap, this paper introduces \textbf{WaveSync}. This framework leverages an LLM to assign importance weights to each word, and these are then combined with Whisper-aligned \cite{radford2023robust} timestamps to construct a Semantic Importance Wave (SIW). Gesture trajectories are then shaped by Dynamic Movement Primitives (DMPs) \cite{Dmp}, which automatically adapt timing and adjust motion style while preserving the original movement. Finally, a Wavefront Optimization module schedules these gestures along the SIW. It applies a crossover rule to reduce temporal conflicts between consecutive gestures and uses a Forward Propagation mechanism to resolve remaining conflicts by inserting natural pauses rather than time-stretching the speech signal. This approach aligns speech emphasis with gesture stroke peaks while respecting the robot's kinematic constraints.
The core contributions of this work are as follows:
\begin{itemize}
\item A hybrid architecture that converts LLM-driven semantic intent into robot motions via DMPs, avoiding the risks associated with directly generating raw joint-angle trajectories.

\item A co-speech gesture synchronization framework based on Wavefront Optimization that performs word-level peak alignment while resolving temporal conflicts under kinematic constraints.

\item A simulation-based evaluation demonstrating improved speech--gesture synchronization, motion smoothness, and perceptual quality across dialogue scenarios.
\end{itemize}

\section{Methodology}
\begin{figure*}[t]
\centering
\includegraphics[width=0.98\linewidth]{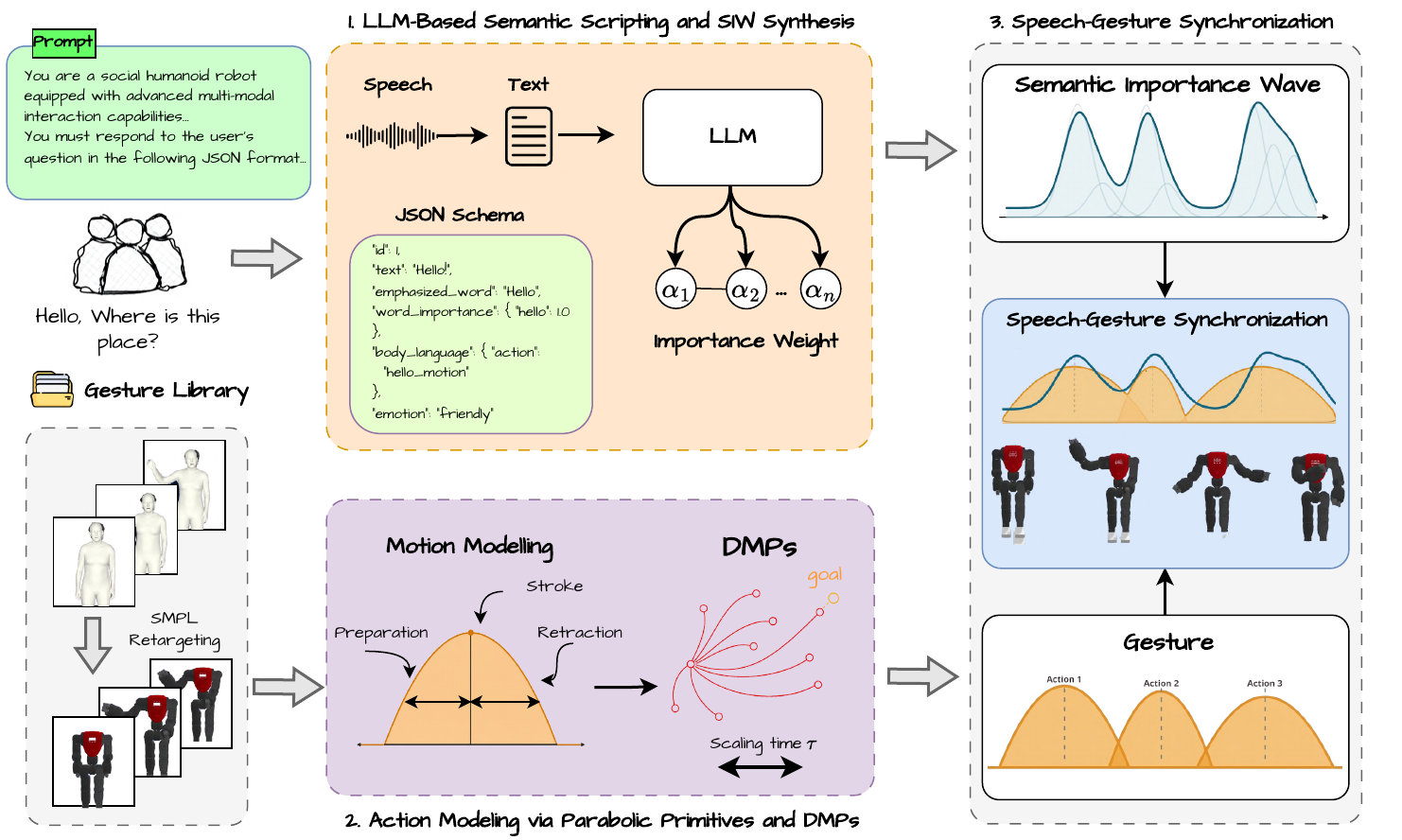}
\caption{The WaveSync pipeline. \textbf{(1) LLM-Based Semantic Scripting and SIW Synthesis:} user speech is processed into a JSON schema with per-word importance weights, which are fused with Whisper-aligned timestamps to form the SIW. \textbf{(2) Action Modeling via Parabolic Primitives and DMPs:} a retrieved gesture is modeled as a parabolic stroke and encoded using a DMP with motion parameters. \textbf{(3) Co-Speech Gesture Synchronization:} the Wavefront Optimization module aligns gesture stroke peaks with SIW peaks while respecting kinematic constraints.}
\label{fig:overall}
\end{figure*}

We present the \textbf{WaveSync} pipeline in three stages, as illustrated in Fig.~\ref{fig:overall}. First, an LLM assigns per-word importance weights, which are combined with Whisper-aligned timestamps to construct the SIW. Second, retrieved gestures are adapted via DMPs for time and amplitude scaling. Finally, a Wavefront Optimization module aligns each gesture stroke peak with its corresponding SIW peak while respecting the robot's kinematic constraints.

\subsection{LLM-Based Semantic Scripting and SIW Synthesis}

First, we transform discrete word-level semantic importance scores into a continuous temporal importance signal that serves as the synchronization target for subsequent modules. Given user input $I$, the LLM generates a response $A$ and decomposes it into prosodic segments $S = \{s_1, \dots, s_n\}$. For each word $w_{j,k}$ in segment $s_j$, the LLM assigns a semantic importance weight $\alpha_{j,k} \in [0,1]$, where values near 1 indicate semantically important words and values near 0 indicate less salient words. Each segment is returned as a structured descriptor:

\begin{equation}
S_j =
\left\{
T_j,\,
\{\alpha_{j,k}\}_{k=1}^{N_j},\,
E_j
\right\},
\end{equation}
where $T_j$ is the emphasized word, $\{\alpha_{j,k}\}_{k=1}^{N_j}$ contains the importance weights of the $N_j$ words in $s_j$, and $E_j$ denotes the motion style parameters.

Since importance weights are defined at the word level rather than in the time domain, a Whisper alignment maps each word $w_{j,k}$ to $[t_{j,k}^{\mathrm{start}},t_{j,k}^{\mathrm{end}}]$, with temporal center:
\begin{equation}
t_{j,k}^c = \frac{t_{j,k}^{\mathrm{start}} + t_{j,k}^{\mathrm{end}}}{2},
\end{equation}
which represents the target timestamp at which the corresponding gesture stroke peak should ideally occur. The discrete pairs $\{(\alpha_{j,k}, t_{j,k}^c)\}$ are then transformed into a continuous signal by superimposing Gaussian kernels:
\begin{equation}
s(t)
=9
\sum_{j=1}^{n}
\sum_{k=1}^{N_j}
\alpha_{j,k}
\exp\left(
-\frac{
\left(t-t_{j,k}^{c}\right)^2
}{
2\sigma^2
}
\right).
\end{equation}
where $\sigma$ controls the temporal spread of each word-level importance peak. The resulting signal is normalized $s(t)\in[0,1]$, and the normalized SIW serves as the synchronization target for the Wavefront Optimization stage.

\subsection{Kinematic Action Modeling via Parabolic Primitives and DMPs}

In parallel, semantic descriptors $S_j$ are translated into expressive, kinematically constrained trajectories through two steps: parabolic gesture modeling and DMP-based modulation. 
Human gestures naturally follow a three-phase structure---preparation, stroke, and retraction \cite{studdert1994hand}. Accordingly, we approximate the gesture stroke profile of the $j$-th gesture as an inverted parabola:
\begin{equation}
P_j(t)=-a_j\bigl(t-t_j^{\mathrm{peak}}\bigr)^2+h_j,\;
t\!\in\![t_j^{\mathrm{start}},t_j^{\mathrm{end}}].
\end{equation}
where $t^{\text{peak}}_{j}$ is the stroke peak timestamp, $h_j$ is the peak magnitude, and $a_j$ is the temporal width of the gesture. 

After that, gestures are selected from a predefined motion library according to the action labels specified in the semantic descriptors. The library contains human demonstrations retargeted to the robot through SMPL-based kinematic mapping \cite{hubo}. Each motion is encoded as a DMP to achieve adaptive timing and adjust motion style.
\begin{equation}
\tau^2 \ddot{x}
=
\alpha_x\bigl(\beta_x(g-x)-\tau\dot{x}\bigr)
+
f(z),
\end{equation}
where $g$ is the goal state, $\alpha_x$ and $\beta_x$ are the DMP convergence gains, $f(z)$ is the nonlinear forcing term, and $\tau$ controls temporal scaling. 

Before Wavefront Optimization, emotion-dependent parameters modify the motion style by adjusting gesture amplitude, anticipation, and spatial offset. The spatial offset shifts the encoded trajectory and consequently its goal state $g$, while the forcing term $f(z)$ controls the gesture shape and amplitude. Small Gaussian perturbations are added to the DMP weights to introduce natural motion variations. Finally, $\tau$ is adjusted for temporal adaptation, allowing the gesture duration $D(\tau)$ to be compressed or expanded while preserving the learned motion pattern.

\subsection{Speech-Gesture Synchronization}
\label{sec:optimization}
\begin{figure}[!t]
\centering
\includegraphics[width=0.9\linewidth]{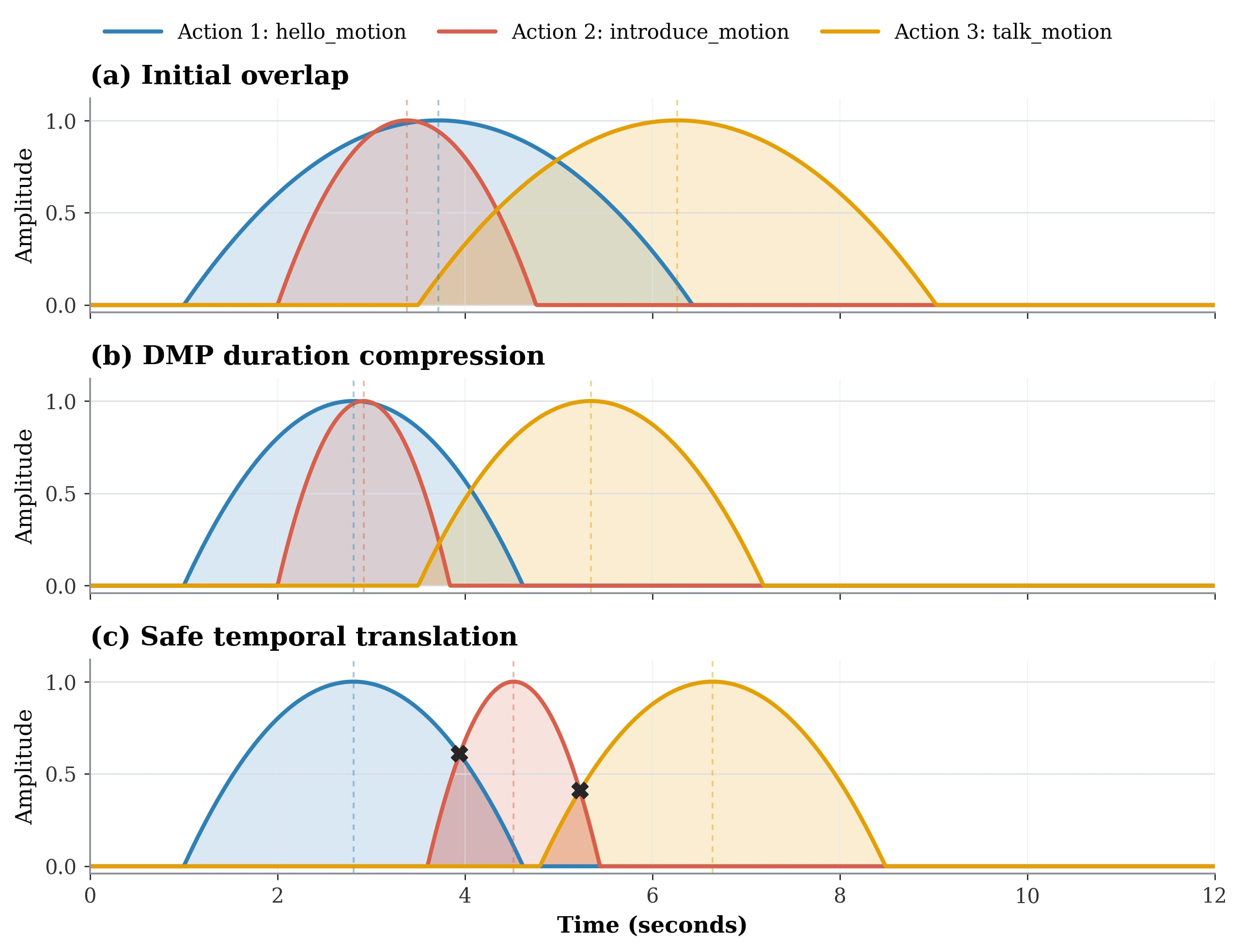}
\caption{Wavefront Optimization: (a) temporal conflict, (b) DMP duration compression, and (c) Forward Propagation under the crossover constraint.}
\label{fig::optimization_process}
\end{figure}

To resolve temporal conflicts that arise when long-duration gestures are assigned to closely spaced semantic peaks in the SIW, we introduce the Wavefront Optimization algorithm. It follows a three-stage hierarchical policy consisting of peak alignment, execution adjustment, and forward propagation, as illustrated in Fig.~\ref{fig::optimization_process}.

First, the peak-alignment mechanism positions each parabolic pulse $P_j$ near the corresponding SIW peak by maximizing their temporal overlap:
\begin{equation}
t_j^{\mathrm{start}*}
=
\arg\max_{t_j^{\mathrm{start}}}
\int_{t_j^{\mathrm{start}}}^{t_j^{\mathrm{end}}}
P_j(t)\,s(t)\,dt,
\label{eq:opt_obj}
\end{equation}
where $t_j^{\mathrm{end}} = t_j^{\mathrm{start}} + D_j(\tau_j)$ and $D_j(\tau_j)$ is the DMP-scaled gesture duration. The time-scaling factor $\tau_j$ is subsequently adjusted when necessary to resolve temporal conflicts.

Peak alignment may cause consecutive gesture profiles $P_j$ and $P_{j+1}$ to overlap because of their different execution durations. Excessive temporal compression may lead to high joint velocities and distorted motion. Therefore, the execution duration of the subsequent gesture is adjusted subject to the lower bound
\begin{equation}
D_{j+1}(\tau_{j+1})
\geq
\rho_D D_{j+1}^{0},
\label{eq}
\end{equation}
where $D_{j+1}^{0}$ denotes the nominal duration of gesture $j+1$, and $\rho_D \in (0,1]$ specifies the maximum allowable compression ratio. In our implementation, $\rho_D = 2/3$, meaning that a gesture is not compressed below two-thirds of its nominal duration. This threshold is selected empirically to avoid excessively rapid execution and preserve the underlying motion profile.

If temporal overlap remains after duration adjustment, a crossover constraint is applied. Let $t_j^{\mathrm{cross}}$ denote the intersection time between two consecutive gesture profiles $P_j$ and $P_{j+1}$. A valid crossover satisfies
\begin{equation}
t_j^{\mathrm{peak}}
\leq
t_j^{\mathrm{cross}}
\leq
t_{j+1}^{\mathrm{peak}},
\qquad
P_{j+1}(t_j^{\mathrm{cross}})
\leq
\rho_C,
\label{eq:crossover}
\end{equation}
where $\rho_C \in (0,1]$ controls the allowable crossover level. In our experiments, we set $\rho_C = 2/3$.

If Eq.~\eqref{eq:crossover} cannot be satisfied after duration compression reaches its lower bound, \emph{Forward Propagation} introduces the minimum temporal delay required to obtain a valid crossover:
\begin{equation}
\Delta_j
=
\min_{\Delta \geq 0}
\left\{
\Delta
\; \middle| \;
C\!\left(P_j, P_{j+1}^{\Delta}\right)=1
\right\},
\end{equation}
where $C(\cdot)$ indicates whether the crossover condition in Eq.~\eqref{eq:crossover} is satisfied, and $P_{j+1}^{\Delta}$ denotes the delayed profile of gesture $j+1$.

The resulting delay is then propagated to subsequent gestures:
\begin{equation}
t_k^{\mathrm{start}}
\leftarrow
t_k^{\mathrm{start}}+\Delta_j,
\qquad \forall\, k>j.
\label{eq:fwd_prop}
\end{equation}

The accumulated delay $\Delta_{\mathrm{total}}=\sum_j \Delta_j$ is incorporated into the TTS schedule as natural inter-segment pauses. The corresponding speech timestamps are updated accordingly to maintain speech--gesture synchronization without time-stretching the synthesized waveform. At gesture boundaries, minimum-jerk interpolation is applied to ensure smooth motion transitions.

\section{Experimental Results}
\label{sec:experiments}

% -----------------------------------------------------------
\subsection{Experimental Setup}
\label{subsec:setup}

All experiments are conducted in PyBullet using a humanoid upper-body model. Speech is synthesized offline with a neural TTS engine, and word-level timestamps are obtained using Whisper alignment. The gesture library contains beat, deictic, and iconic motions retargeted from human demonstrations via SMPL-based kinematic mapping. Each clip is semantically tagged for gesture retrieval.

\textbf{Dialog Scenarios.}
Five dialogue scenarios (S1--S5) covering diverse communicative contexts are evaluated, including greeting, casual conversation, description, warning, and thinking.

\begin{table}[ht]
  \centering
  \caption{Dialogue scenarios and emphasized words used in the evaluation (14 gesture events in total).}
  \label{tab:utterances}
  \scriptsize
  \setlength{\tabcolsep}{4pt}
  \renewcommand{\arraystretch}{1.08}
  \begin{tabularx}{\linewidth}{c | l | X}
    \hline
    \textbf{ID} & \textbf{Function} & \textbf{Dialogue} \\
    \hline
    S1 & Greeting &
    \textcolor{red}{Hello} everyone, \textcolor{red}{welcome} to \textcolor{red}{our} research space. \\
    \hline
    S2 & Casual &
    This \textcolor{red}{issue} may seem \textcolor{red}{small}, but if \textcolor{red}{overlooked}, it leaves \textcolor{red}{significant} consequences. \\
    \hline
    S3 & Descriptive &
    When mentioning elephants, the first impression is they are \textcolor{red}{enormous}: they reach very \textcolor{red}{high} and their bodies are extremely \textcolor{red}{large}. \\
    \hline
    S4 & Warning &
    \textcolor{red}{Attention}! You absolutely \textcolor{red}{must not} step over this \textcolor{red}{safety line}. \\
    \hline
    S5 & Thinking &
    \textcolor{red}{Wait} a moment\ldots\ Let's see what we can do. \\
    \hline
  \end{tabularx}
\end{table}

\textbf{Ablation Variants.}
Three ablated configurations are evaluated to isolate the contribution of the main WaveSync components:

\begin{itemize}
  \item \textbf{w/o~SIW}: Gestures are triggered uniformly at word boundaries without semantic peak-based alignment.
  \item \textbf{w/o~DMP}: Pre-recorded keyframes are executed without DMP-based trajectory shaping or temporal scaling.
  \item \textbf{w/o~Wavefront}: The hierarchical scheduling mechanism is disabled; temporally conflicting gestures are directly transitioned without duration adjustment or forward propagation.
\end{itemize}

% -----------------------------------------------------------
\subsection{Objective Evaluation}
\label{subsec:metrics}

\subsubsection{Metrics}
 
\textbf{Human-based Sync (HbS).}
To obtain the human reference for synchronization evaluation, ten participants independently annotated the desired gesture-peak timing by pressing the spacebar at the moment they perceived a gesture stroke should coincide with speech emphasis. The annotations for each gesture event $i$ were aggregated into a trimmed-mean consensus timestamp $\bar{T}_{i}^{*}$, which serves as the reference for computing PAM and HbS.

For each gesture event $i$, the Phase Alignment Metric (PAM) measures the absolute temporal deviation between the predicted gesture-peak timestamp $\hat{T}_{i}^{(m)}$ and the human consensus reference $\bar{T}_{i}^{*}$:
\begin{equation}
\mathrm{PAM}_{i}^{(m)}
=
\left|
\hat{T}_{i}^{(m)}
-
\bar{T}_{i}^{*}
\right|.
\label{eq:pam_event}
\end{equation}

The event-level HbS score normalizes PAM using a tolerance parameter $\tau = 1\,\mathrm{s}$~\cite{giorgolo2008perception}, providing a bounded measure of temporal agreement between the generated gesture peak and the human reference:
\begin{equation}
\mathrm{HbS}_{i}^{(m)}
=
1
-
\frac{
\min\left(
\mathrm{PAM}_{i}^{(m)},
\tau
\right)
}{
\tau
}.
\label{eq:hbs_event}
\end{equation}

The overall HbS score is then averaged across all gesture events in $\mathcal{S}$:
\begin{equation}
\mathrm{HbS}^{(m)}
=
\frac{1}{|\mathcal{S}|}
\sum_{i \in \mathcal{S}}
\mathrm{HbS}_{i}^{(m)}.
\label{eq:hbs_final}
\end{equation}

A score closer to $1$ indicates stronger temporal agreement. Errors larger than $\tau$ are assigned a score of zero. The tolerance is used as a conservative normalization bound rather than as a strict perceptual synchrony threshold. Because clamping is performed at the event level before averaging, $\mathrm{HbS}$ is not necessarily equal to $1-\mathrm{PAM}/\tau$ when individual events exceed the tolerance.
 
\textbf{Joint Jerk Magnitude.}
Motion smoothness is quantified using the $\ell_2$-norm of the third temporal derivative of the joint-angle vector:
\begin{equation}
  J(t) = \left\|\frac{d^3\mathbf{q}(t)}{dt^3}\right\|_2 = \left\|\dot{\mathbf{a}}(t)\right\|_2,
  \label{eq:jerk}
\end{equation}

where $\mathbf{q}(t)$ denotes the vector of commanded joint angles. Lower jerk indicates smoother temporal evolution of the joint trajectories.
 
\subsubsection{Results}

\textbf{Synchronization Accuracy (HbS / PAM).}
Table~\ref{tab:hbspam} reports the temporal alignment performance of WaveSync across five dialogue scenarios. The method achieves an unweighted scenario-average PAM of $838.0\,\mathrm{ms}$ and HbS of $0.307$, while the event-weighted HbS over all 14 gesture events is $0.289$; this difference results from scenario-level versus event-level averaging. S3 achieves the best alignment, whereas S2 shows the largest temporal deviation. Overall, the results indicate that WaveSync maintains meaningful speech--gesture correspondence across the evaluated scenarios, although the alignment is not always tight.

\begin{table}[ht]
  \centering
  \caption{Temporal alignment performance of WaveSync across five dialogue scenarios. The final row reports the unweighted average across scenarios.}
  \label{tab:hbspam}
  \scriptsize
  \setlength{\tabcolsep}{4pt}
  \renewcommand{\arraystretch}{1.08}
  \begin{tabular}{c c c c c}
    \hline
    \textbf{Scen.} &
    \textbf{Events} &
    \textbf{Mean PAM} $\downarrow$ &
    \textbf{Median PAM} $\downarrow$ &
    \textbf{HbS} $\uparrow$ \\
    \hline
    S1 & 3 & 839.3 ms  & 952.3 ms  & 0.233 \\
    S2 & 4 & 1261.4 ms & 1244.3 ms & 0.113 \\
    S3 & 3 & 466.1 ms  & 408.1 ms  & 0.534 \\
    S4 & 3 & 960.2 ms  & 609.8 ms  & 0.319 \\
    S5 & 1 & 663.0 ms  & 663.0 ms  & 0.337 \\
    \hline
    \textbf{Scenario Avg.} &
    14 &
    \textbf{838.0 ms} &
    \textbf{775.5 ms} &
    \textbf{0.307} \\
    \hline
  \end{tabular}
\end{table}

\begin{figure*}[t]
  \centering
  \includegraphics[width=0.83\linewidth]{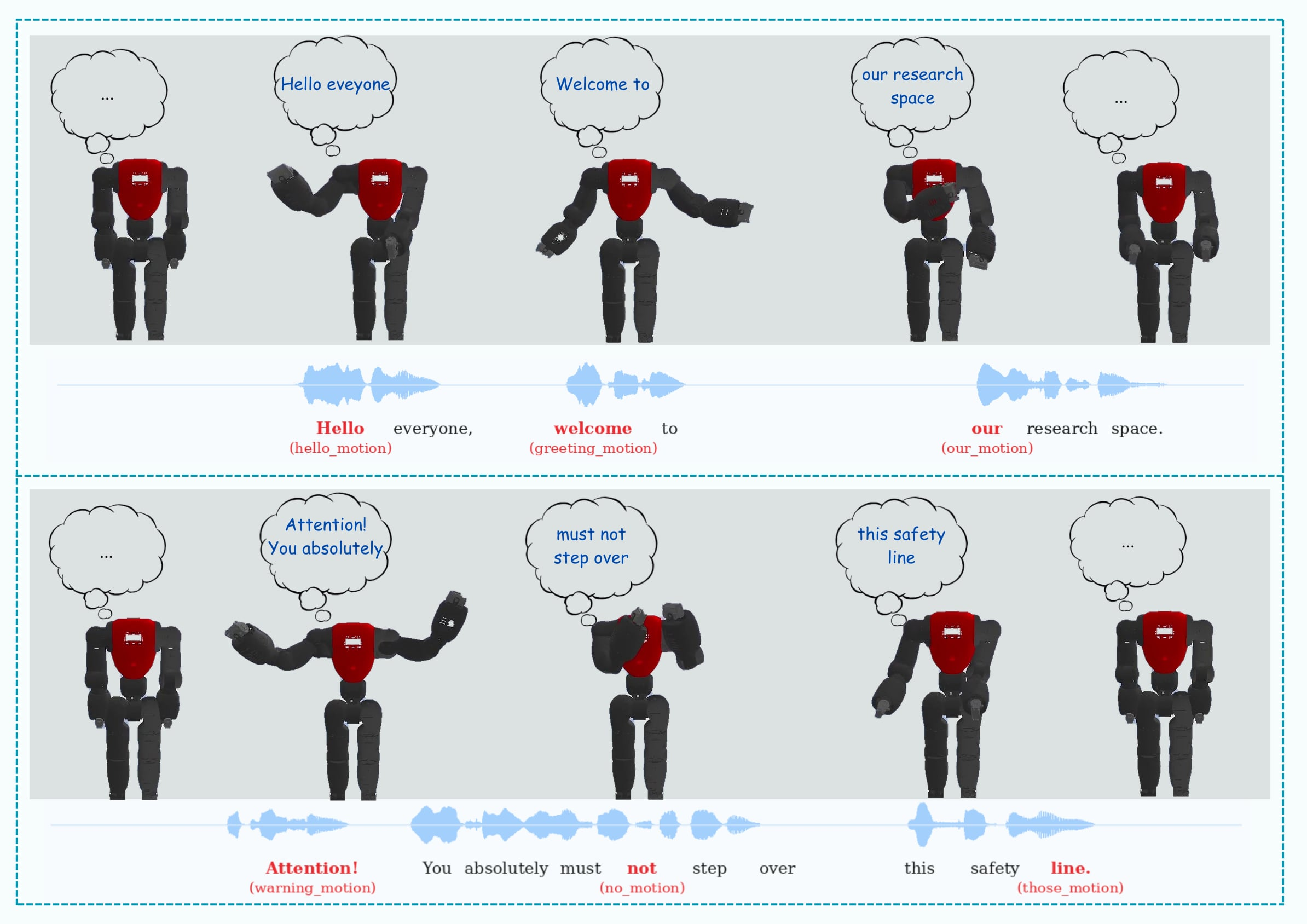}
  \caption{Gesture sequences for S1 (top) and S4 (bottom). Highlighted words indicate SIW emphasis peaks, and labels denote retrieved gestures.}
  \label{fig:result_demo}
\end{figure*}

Figure~\ref{fig:result_demo} presents qualitative results for S1 and S4, showing the final synchronized gesture sequences generated by WaveSync during execution.

\begin{figure}[t]
  \centering
  \includegraphics[width=\linewidth]{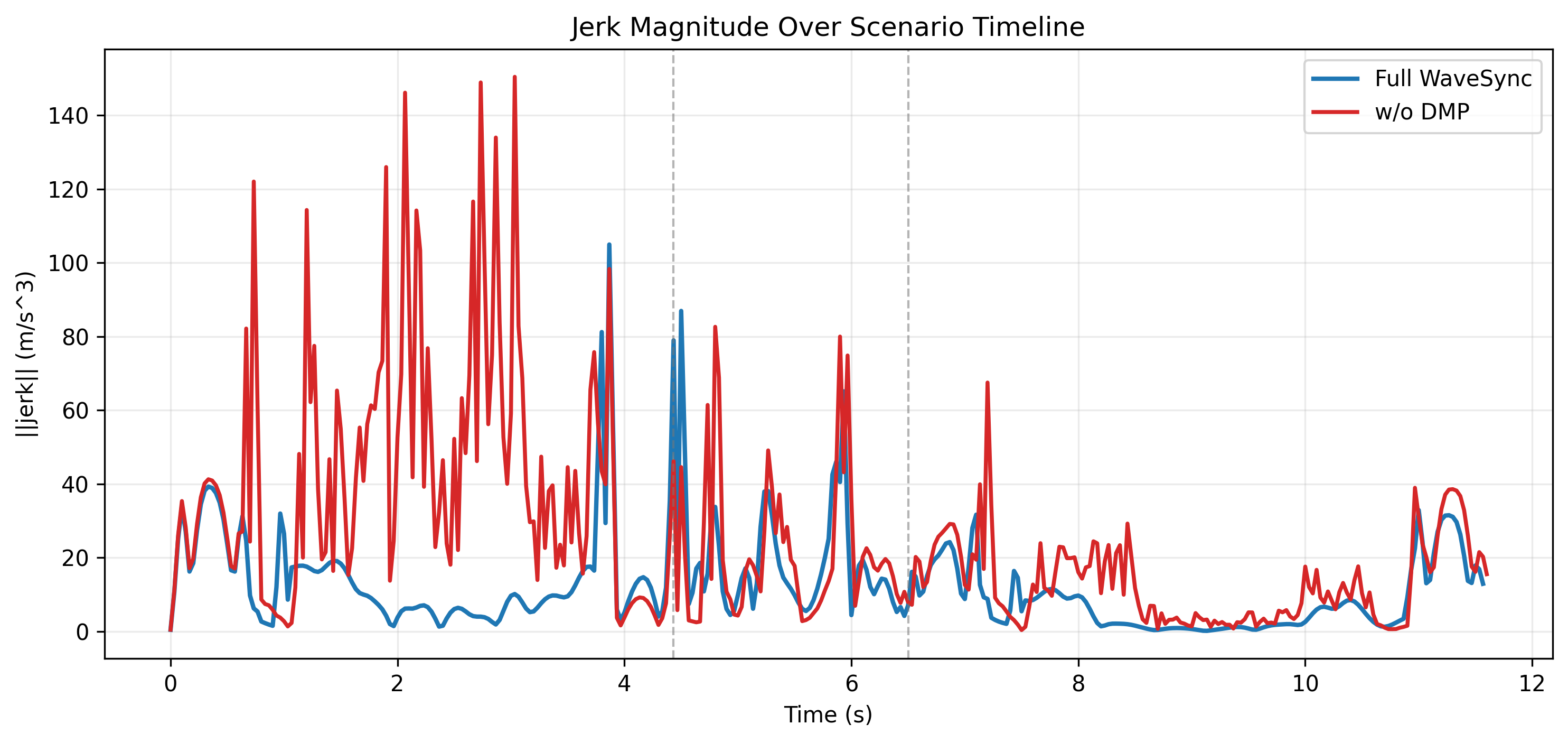}
  \caption{Joint jerk magnitude $J(t)$ in Scenario~1. Lower jerk indicates smoother joint trajectories.}
  \label{fig:jerk}
\end{figure}

\textbf{Trajectory Smoothness (Jerk).}
Figure~\ref{fig:jerk} compares joint jerk magnitude for WaveSync and \textit{w/o~DMP} in Scenario~1. WaveSync exhibits lower and more stable jerk, indicating that DMP-based trajectory shaping reduces abrupt joint variations and improves motion smoothness.

\subsection{Subjective Evaluation}
\label{subsec:subjective}

A separate group of ten participants rated WaveSync and the three ablated variants in randomized order using a 5-point Likert scale for Motion Smoothness (MS), Temporal Synchronization (TS), Semantic Appropriateness (SA), and Overall Naturalness (ON).

\begin{figure}[t]
  \centering
  \includegraphics[width=0.48\textwidth]{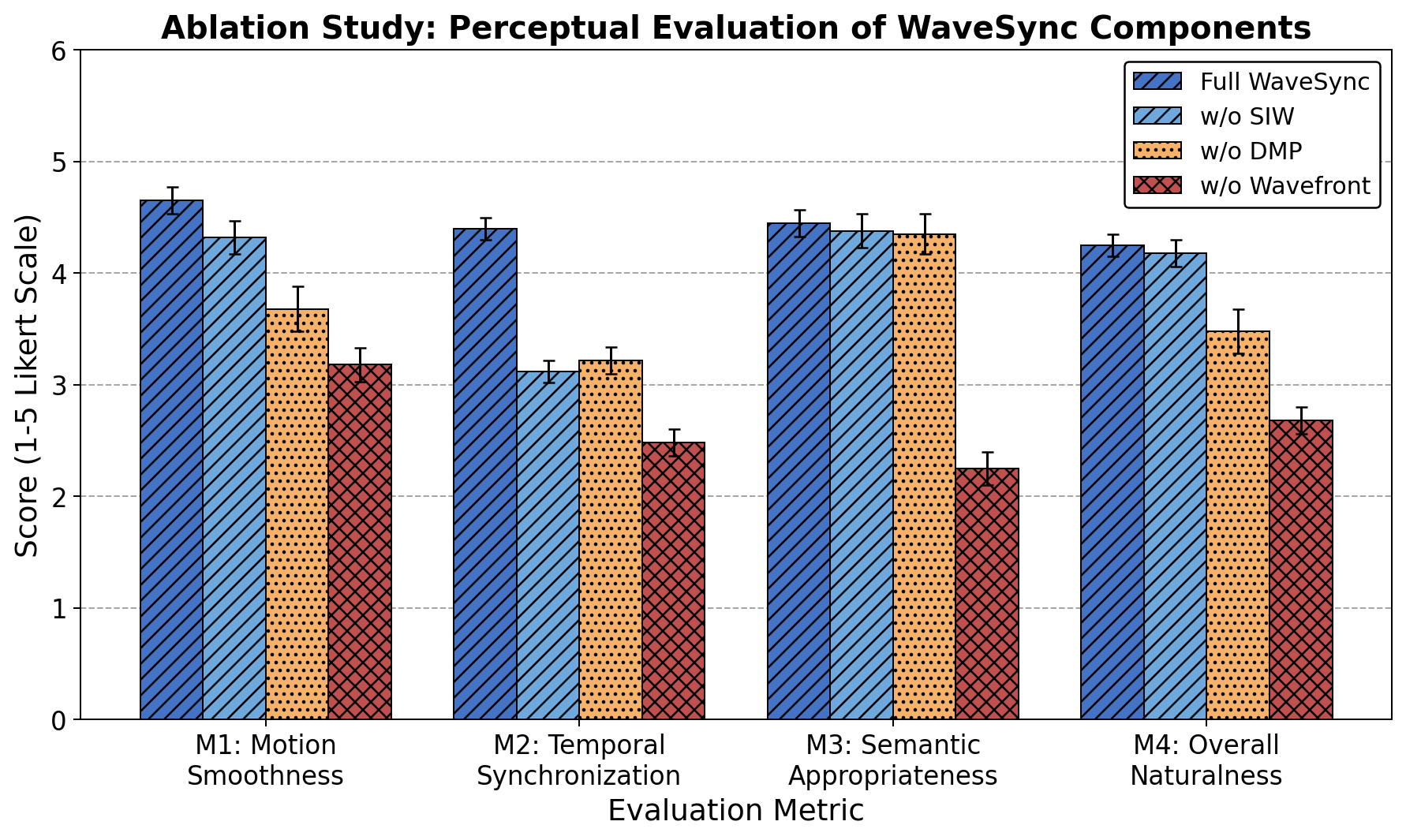}
  \caption{Perceptual Likert scores (MS, TS, SA, ON) for WaveSync and three ablated variants. Error bars show $\pm$1\,SD.}
  \label{fig:ablation}
\end{figure}

Figure~\ref{fig:ablation} presents the perceptual ablation results. WaveSync achieves the highest mean score across all four dimensions. Removing DMP-based shaping reduces Motion Smoothness (MS) to approximately $3.7$, consistent with the higher jerk observed in the quantitative trajectory analysis. Removing the Wavefront scheduler further reduces MS to approximately $3.2$, indicating that unresolved temporal conflicts degrade motion smoothness. The \textit{w/o~SIW} variant shows the largest decrease in Temporal Synchronization (TS), supporting the role of the SIW as the primary semantic timing signal. Semantic Appropriateness (SA) is lowest for \textit{w/o~Wavefront} ($\sim2.2$), suggesting that unresolved gesture conflicts can disrupt the perceived correspondence between speech and motion. Overall, the results indicate complementary contributions from SIW-based timing, DMP-based motion shaping, and Wavefront Optimization.

\section{Conclusion}
\label{sec:conclusion}

We presented WaveSync, a framework for synchronizing expressive robot gestures with speech under semantic and kinematic constraints. By combining Semantic Importance Weights, DMP-based trajectory adaptation, and Wavefront Optimization, the system aligns gesture stroke peaks with speech-emphasis targets while resolving temporal conflicts through duration adjustment and forward propagation. Experiments across five dialogue scenarios demonstrated within-tolerance speech--gesture alignment and smoother motion trajectories, while the perceptual ablation study showed that WaveSync achieved the highest ratings across all evaluated dimensions.

The results further indicate that SIW-based timing, DMP-based motion shaping, and Wavefront Optimization provide complementary contributions to the final behavior. However, the current system relies on a fixed gesture library, which limits its coverage of unconstrained dialogue. Future work will explore automatic gesture synthesis and evaluate WaveSync on real humanoid robot hardware.

%===============================================================================
\section*{Acknowledgment}
This work was supported by JST SPRING, Japan Grant Number JPMJSP2102.

\bibliographystyle{IEEEtran}
\bibliography{references}  % .bib

\newpage
\end{document}